\documentclass[a4paper]{article}

\usepackage{xcolor}
\usepackage[noadjust]{cite}

\usepackage[pdftex]{graphicx}
\usepackage{fullpage}
\usepackage[caption=false,font=footnotesize]{subfig}

\usepackage{amsmath}
\usepackage{url}

\usepackage{float,amssymb}

\makeatletter
\def\blfootnote{\xdef\@thefnmark{}\@footnotetext}
\makeatother

\begin{document}

\title{Efficient Unsupervised Learning\\ for Plankton Images}


\author{%
{\bf Paolo Didier Alfano}$^{1}$\hfill \hspace{14em}{\small\texttt{paolodidier.alfano@edu.unige.it}}\\
 {\small \it MaLGa - DIBRIS, University of Genova, Italy \hfill \hspace{12em}}\\
 \vspace{-1em}
 \and
 {\bf Marco Rando}$^{1}$\hfill \hspace{20em}\small \texttt{marco.rando@edu.unige.it}\\
 {\small \it MaLGa - DIBRIS, University of Genova, Italy \hfill \hspace{12em}}\\
 \vspace{-1em}
 \and
 {\bf Marco Letizia} \hfill \hspace{19em}{\small\texttt{marco.letizia@edu.unige.it}}\\
 {\small \it MaLGa - DIBRIS, University of Genova, Italy \hfill \hspace{12em}}\\  
 {\small \it INFN - Sez. di Genova, Genova, Italy \hfill \hspace{12em}}\\
 \vspace{-1em}
 \and
 {\bf Francesca Odone} \hfill \hspace{18.3em}{\small\texttt{francesca.odone@unige.it}}\\
 {\small \it MaLGa - DIBRIS, University of Genova, Italy \hfill \hspace{12em}}
 \and
 {\bf Lorenzo Rosasco} \hfill \hspace{18.3em}{\small\texttt{lorenzo.rosasco@unige.it}}\\
 {\small \it MaLGa - DIBRIS, University of Genova, Italy \hfill \hspace{12em}}\\
 {\small \it Istituto Italiano di Tecnologia, Genova, Italy \hfill \hspace{12em}}\\
 {\small \it CBMM - MIT, Cambridge, MA, USA \hfill \hspace{12em}}
  \and
 {\bf Vito Paolo Pastore} \hfill \hspace{15.5em}{\small\texttt{vito.paolo.pastore@unige.it}}\\
 {\small \it MaLGa - DIBRIS, University of Genova, Italy \hfill \hspace{12em}}\\
}

\maketitle

\begin{abstract}
Monitoring plankton populations in situ is fundamental to preserve the aquatic ecosystem. Plankton microorganisms are in fact susceptible of minor environmental perturbations, that can reflect into consequent morphological and dynamical modifications.  Nowadays, the availability of advanced automatic or semi-automatic acquisition systems has been allowing the production of an increasingly large amount of plankton image data. The adoption of machine learning algorithms to classify such data may be affected by the significant cost of manual annotation, due to both the huge quantity of acquired data and the numerosity of plankton species. 
To address these challenges, we propose an efficient unsupervised learning pipeline to provide accurate classification of plankton microorganisms. 
We build a set of image descriptors exploiting a two-step procedure. First, a Variational Autoencoder (VAE) is trained on features extracted by a pre-trained neural network. We then use the learnt latent space as image descriptor for clustering.
We compare our method with state-of-the-art unsupervised approaches, where a set of pre-defined hand-crafted features is used for clustering of plankton images. The proposed pipeline outperforms the benchmark algorithms for all the plankton datasets included in our analysis, providing better image embedding properties.    

\end{abstract}

\maketitle
\footnotetext[1]{These authors contributed equally to this work.}
\blfootnote{\textcopyright 2022 IEEE. Personal use of this material is permitted. Permission from IEEE must be obtained for all other uses, in any current or future media, including reprinting/republishing this material for advertising or promotional purposes, creating new collective works, for resale or redistribution to servers or lists, or reuse of any copyrighted component of this work in other works.}

\section{Introduction}\label{sec:intro}

Plankton is a collection of aquatic microorganisms floating passively in the water. It plays a big role in the marine ecosystem. Plankton is indeed at the basis of the aquatic food chain, with phytoplankton being estimated to have produced approximately 50\% of the total oxygen in our atmosphere \cite{behrenfeld2001biospheric}. Recently, it has been proved that local or global perturbations of the aquatic environment, either natural or man-caused, are profoundly impacting both the composition and dynamics of plankton populations \cite{Boyce}. 
As stated in \cite{pastore2019establishing}, plankton microorganisms react to even minimal changes in the environment with morphological and dynamical modifications, so they can be regarded as biosensors, reflecting the overall health of the oceans.
Thus, detecting and studying plankton population in situ, is paramount to protect marine ecosystems \cite{lumini2019deep}.
Recently, the development of advanced tools for automatic high-throughput in situ microscopy allowed real-time observation of plankton species \cite{Fossum,Pastore2020}. Such systems acquire a huge amount of image data for which manual identification is impractical \cite{sosik2007automated}.   
Hence, machine learning has nowadays become one of the most studied approaches for the characterization of plankton data \cite{culverhouse1994automatic,hu2005automatic,4129463,https://doi.org/10.4319/lom.2007.5.204,benfield2007rapid,Schulze2013,SCHMID2016129,Zheng2017,pastore2019establishing,biswas2019high}. In particular, there has been a surge in interest towards models based on artificial neural networks (ANNs), due to their successes in big data problems and their high expressive power, specifically in the form of convolutional neural networks (CNNs) \cite{dieleman2016exploiting,7485680,7404463,rathi2017underwater,https://doi.org/10.1002/lom3.10285}. 
A central element in the process of characterizing plankton data is feature selection. The two main approaches are represented by hand-engineered features, such as geometric or Fourier features \cite{ZHAO20101853,Zheng2017}, and deep features, typically based on deep CNNs \cite{du_toit_enhanced_2021,7485680}.
Recent works \cite{lumini2019deep,orenstein2017transfer,lee2016plankton} have also explored the possibility to use pre-training and transfer learning (both in- and out-of-domain) to enhance the expressivity of the models and alleviate the computational cost associated with the training of deep CNNs. In \cite{ellen2019improving}, the authors demonstrate how context metadata, such as temperature, location and salinity, improves the performance of classifiers.
The literature is however largely dedicated to supervised approaches, for which accurate models can be obtained at the high cost of providing manually annotated data. Beside the development of solutions aimed at automatizing the labeling procedures \cite{Hughes164087,schroder2020morphocluster}, research on unsupervised models \cite{Pastore2020} is crucial to avoid bottlenecks in our ability to process information \cite{10.1093/icesjms/fsz057}.

In this work we propose a new unsupervised approach for the characterization of plankton images. There are more than 4000 existing plankton species, and many of them are very similar, from a morphological point of view, making this problem very challenging\cite{anomalydet}. We leverage neural network models that are pre-trained on large general purpose datasets, such as ImageNet, to extract expressive feature maps in an efficient way, without fine-tuning. We then use an encoder-decoder network architecture to perform dimensionality reduction, producing low-dimensional embedded features that can then be fed to a clustering algorithm. We tested our pipeline on three plankton datasets with different characteristics, showing that our results surpass state-of-the-art approaches where hand-crafted features (i.e., geometric and texture descriptors) are engineered and used for clustering. Specifically, the main contributions of this work are: (1) A new VAE-based methodology for efficient unsupervised learning, leveraging the VAE low dimensional and regular latent space. A key novelty consists in using pre-trained features as informative input to the VAE, to obtain high quality descriptive features.  Our findings confirm that VAE can be used not only for generation purposes, but also to discover patterns from data in an unsupervised fashion. (2) We show that the high quality low-dimensional embedding produced by our approach, appears to be informative and allows the effective usage of unsupervised machine learning algorithms, such as k-means or Gaussian mixture models, taming the curse of dimensionality.
(3) We further show that the produced embedding also leads to high accuracy when used as compressed input to supervised models, that can be trained quickly and on limited resources.
The proposed pipeline is efficient and ideal for in situ analysis as well as for off-line investigations of plankton data.

The paper is organized as follows. In Section \ref{sec:method}, we outline and review the proposed pipeline in details and some common architectures that are used in our study. In Section \ref{sec:experiments}, we describe the datasets included in our analysis, present the experimental setup and discuss our results. In Section \ref{sec:conclusions} we lay out our conclusions and discuss future developments.

\section{Method}\label{sec:method}
In this section, we describe the proposed pipeline and briefly review some common neural network architectures that we utilized in our work. 
\subsection{Pipeline}\label{subsec:pipeline}
\begin{figure*}[!ht]
	\begin{center}
		\includegraphics[width=0.9\textwidth]{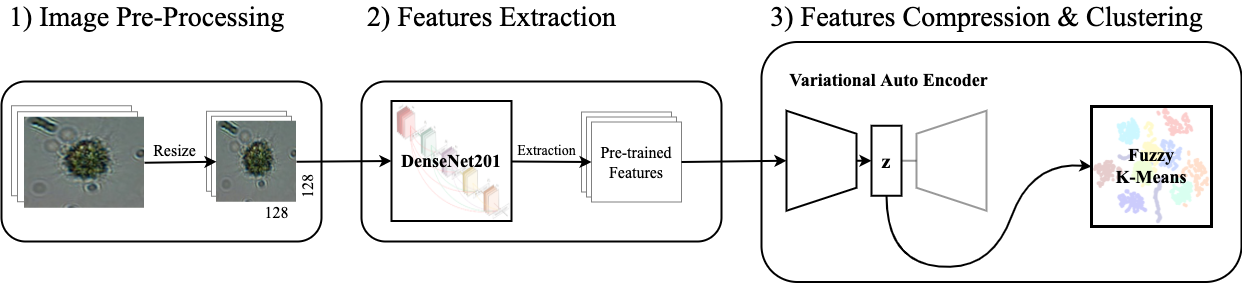}
		\caption{The proposed pipeline made by three steps: Image Pre-processing, Features Extraction, Features compression and Clustering}
		\label{fig:Pipeline}
	\end{center}
\end{figure*}
We first describe our pipeline, which is schematically represented in Figure \ref{fig:Pipeline}. It includes the following steps:
\begin{enumerate}
    \item Pre-processing: images are resized and normalized to be compatible with the neural network used in the next step.
    \item Features extraction: pre-processed images are given as input to a deep neural network pre-trained on the ImageNet dataset \cite{deng2009imagenet}. The output of this step is a high-dimensional feature vector for each image.
    \item Dimensionality reduction and clustering: feature vectors are reshaped and used as input to train a convolutional Variational Auto Encoder (VAE). The learnt latent space is exploited to map input features into a low-dimensional embedding that is finally fed to a clustering algorithm. 
    
\end{enumerate}

In the next paragraphs we provide a general description of the methods involved in each of the steps of the proposed pipeline. In \textit{Experiment Details} (section \ref{subsec:exp_details}) we report specific details about the pipeline implementation. 

\subsection{Features extraction}\label{subsec:background}
\paragraph*{Convolutional Neural Networks} CNNs are a class of feed-forward ANNs commonly utilized to process grid-like inputs, such as images. They are characterized by the convolution operation, which replaces the usual matrix multiplication of fully connected neural networks \cite{Goodfellow-et-al-2016}. The convolution is applied to the input via a kernel matrix and the output is a feature map. 
From a mathematical point of view we can see a CNN in the following way
\begin{equation}\label{cnn}
\Phi=\underbrace{\Phi_{C}\circ \ldots  \circ \Phi_{1}}_{\text{Convolutional layers}},
\end{equation}
where $\Phi_{\ell}$ represents the $\ell$-th layer of the network. In particular, each layer in the feature map usually includes linear and non-linear operations
\begin{equation}
\Phi_\ell(z) = (\sigma(W_\ell^1 \star z), \dots, \sigma(W_\ell^{u_\ell} \star z)),
\end{equation}
where $\star$ represents the convolution operation, $u_\ell$ the number of hidden units in layer $\ell$, $W_\ell$ is a $u_{\ell-1}$ times $u_{\ell}$ matrix, $z$ the input from the previous layer, and where
\begin{equation}
\sigma(a)= \rho( \text{ReLU}(a^1), \dots ,\text{ReLU}(a^{u_\ell}) ),
\end{equation}
with $a\in \mathbb R ^{{u}_\ell}$ and $\rho$ is the max-pooling operation.

\paragraph*{Pre-training} In a typical training scenario, the weights of a model are adjusted, starting from a random initialization, to optimize a measure that is determined by the task at hand, such as accuracy for classification, over a validation set. However, the quality of the results can be impacted by the scarcity of data or by the computational cost of training, especially in the context of deep learning.
One common solution to this kind of problem is represented by the use of \textit{pre-trained} models. The goal is to exploit large state-of-the-art models, the weights of which have been already optimized for a similar task. Usually, the pre-training is performed on a large and general purpose dataset, such as the ImageNet dataset \cite{deng2009imagenet}. A pre-trained model can then be used as is, possibly embedded with frozen weights in a larger model, or by fine tuning it, hence further optimizing its weights on a specific task. In our study, we use a pre-trained model, DenseNet201 \cite{huang2018densely}, as a feature extractor, without fine-tuning. This is a cheap and efficient way to produce better representations of our input images, as it only involves an inference step.

\subsection{Dimensionality reduction and clustering}\label{subsec:dim_Red}
\begin{figure}[H]
	\centering
	\includegraphics[width=\linewidth]{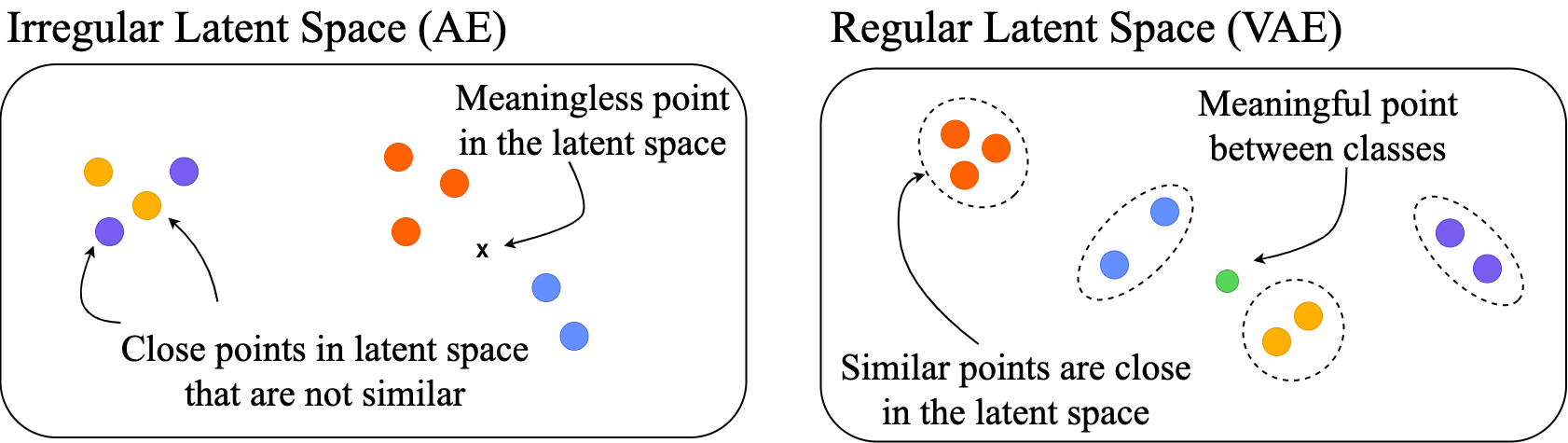}
	\caption{Differences between an irregular and regular latent space.}\label{fig:latspace}
\end{figure}

\begin{figure}[H]
	\centering
	\includegraphics[width=\linewidth]{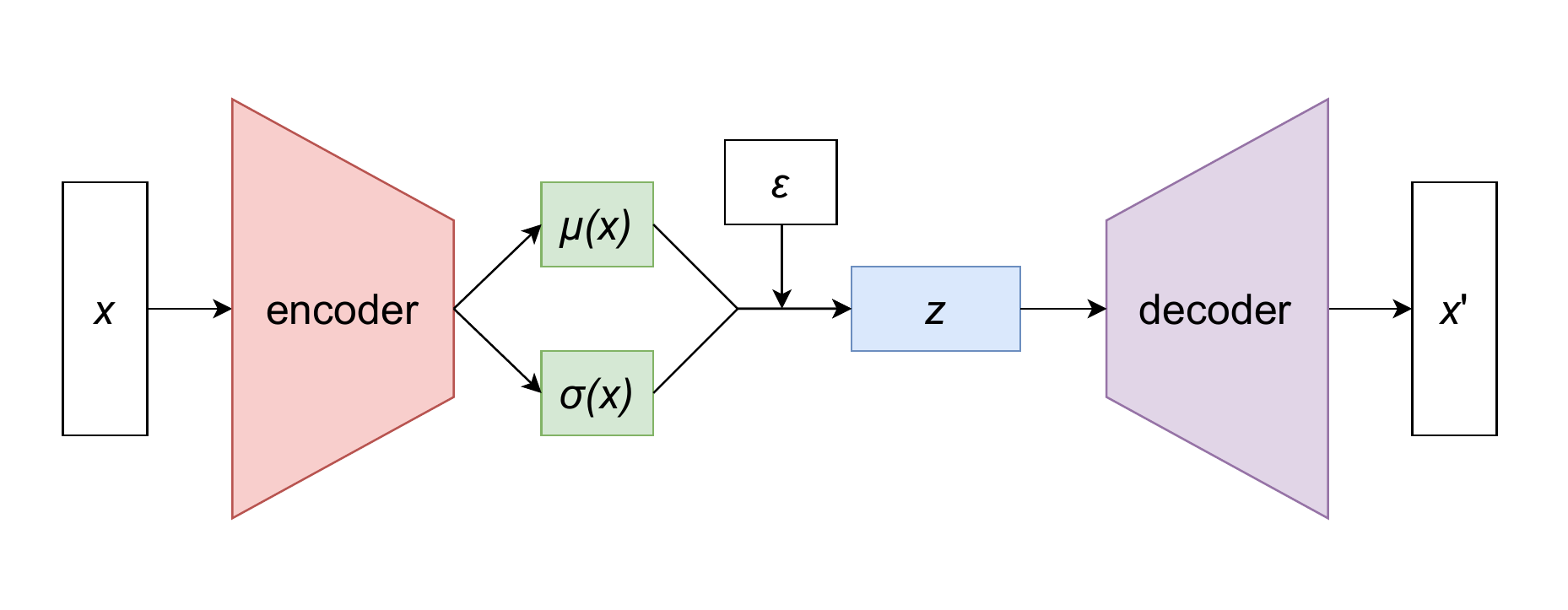}
	\caption{The scheme of a variational autoencoder.}\label{fig:vae}
\end{figure}

\paragraph*{Autoencoders} AEs \cite{kramer1991nonlinear,hinton2006reducing}
are unsupervised machine learning models trained to reproduce its inputs while learning a lower dimensional latent representation. 
Concretely, AEs address this task by learning an \textit{encoder} function $e$ which maps input data $x \in \mathbb{R}^d$ in a latent space $Z \subseteq \mathbb{R}^p$ with $0 < p < d$ and a \textit{decoder} function $d$ which maps a latent representation $z = e(x)$ back to the original input $x$. These maps are typically parametrized by neural networks. Let us denote the output with $x^\prime = d(e(x))$, AEs are then trained to minimize the \textit{reconstruction loss}  
\begin{equation}
	l(x, x^\prime)  = \left\lVert x - x^\prime \right\rVert^2.
\end{equation}
They can be interpreted as non-linear dimensionality reduction models. 
These models provide good performances in data compression, but they cannot be used to generate new instances since they do not impose any structure on the latent space, see Figure \ref{fig:latspace}.
Moreover, minimizing the reconstruction loss without any kind of regularization might lead to overfitting.
\paragraph*{Variational Autoencoder.} VAEs \cite{kingma2014autoencoding} can be seen as a way to improve the structure of the latent space by encoding the input into a multivariate latent distribution. 
More concretely, the encoder maps each input to a multivariate Gaussian distribution $q(z|x)$ with mean and variance parametrized by neural networks
\begin{equation}
q(z|x)=\mathcal{G}(\mu(x),\sigma(x)).
\end{equation}
A sample from the latent distribution $z\sim q(z|x)$ is then decoded and the resulting output $x'=d(z)$ is used to compute the error. The loss function includes a reconstruction terms, as in the AE loss, and a regularization term given by the Kullback-Leibler divergence between the encoded distribution and the prior on the latent representation, assumed normal $p(z) = \mathcal{N}(0, I)$. The total loss reads as
\begin{equation}
    l(x,x')=\left\lVert x - x^\prime \right\rVert^2 + D_{KL}(\mathcal{G}(\mu(x),\sigma(x)), \mathcal{N}(0, I)).
\end{equation}
In order to allow error backpropagation through the encoder, the sampling step is expressed via the \textit{reparametrization trick}:
\begin{equation}
	z = \mu(x) + \epsilon \sigma(x) \qquad \epsilon \sim \mathcal{N}(0, I).
\end{equation}
Ultimately, it is the regularization term that forces the model to learn meaningful latent space representations. Without it, the VAE would try to simply reconstruct the input as closely as possible, for instance, by mapping each input to a delta distribution in the latent space, similarly to an AE, see Figure \ref{fig:latspace}. See Figure \ref{fig:vae} for a schematic representation of a VAE architecture.
A more detailed and formal discussion can be found in \cite{kingma2014autoencoding,kingma2019introduction}.
\paragraph*{Clustering accuracy} \label{purity}The quality of a clustering algorithm can be evaluated using purity, defined as
\begin{equation}
\text{purity}(\Omega, C)=\frac{1}{N}\sum_k\max_j|w_k\cap c_j|,
\end{equation}
where $\Omega=\{w_1,...,w_k\}$ is the set of clusters and $C=\{c_1,...,c_j\}$ is the set of ground-truth classes. Every cluster is then associated with the most represented class.
A purity value of one means that clusters perfectly overlap with the ground truth. Purity decreases when samples from the same class are spread among different clusters, or separate clusters are assigned to the same class. In plankton image analysis, several species have nearly indistinguishable morphological features. Thus, the clustering algorithm can potentially group such species into the same cluster. In our results, we refer to the number of ground truth species overlapping with the same clusters as \textit{number of overlaps}. A number of overlaps equal to zero means that the correspondence between ground truth species and cluster is 1:1, meaning that each cluster represents a unique class (i.e., there is no two or more species that mostly overlap with the same cluster).
We adopted the customized purity implementation described in \cite{Pastore2020}, where the number of overlaps is introduced and used to evaluate clustering performances together with the purity.

\section{Experiments}\label{sec:experiments}

\begin{figure*}[!ht]
\begin{center}
\subfloat[Lensless dataset.]{\includegraphics[width=0.9\textwidth]{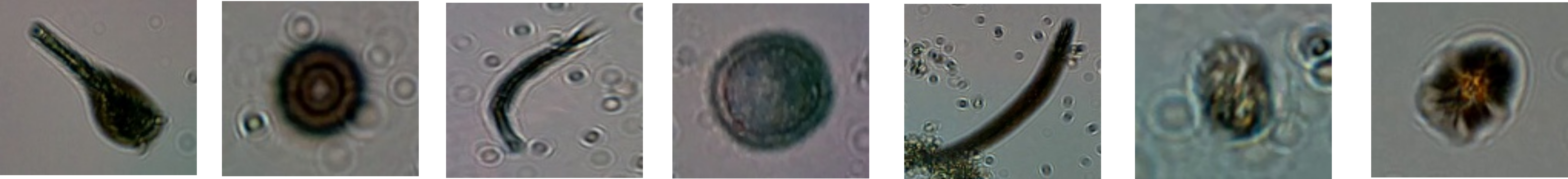}
\label{fig:lenslessdata}}
\hfil
\subfloat[WHOI datasets.]{\includegraphics[width=0.9\textwidth]{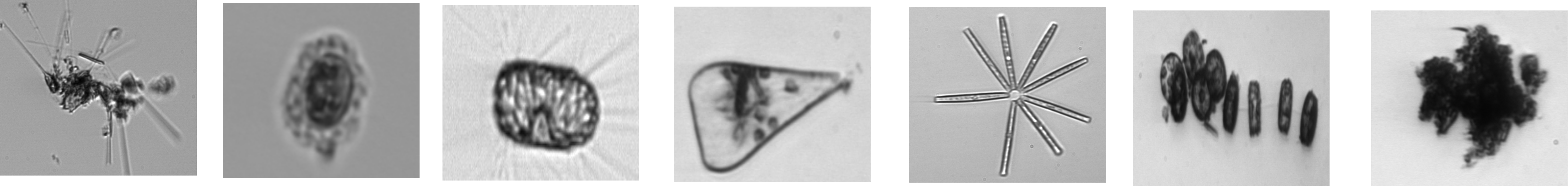}
\label{fig:whoi}}
\caption{Sample images from seven different classes included in the datasets considered for our analysis.} \label{fig:datasets}
\end{center}
\end{figure*}

\subsection{Datasets}\label{subsec:datasets}
We considered three datasets for the evaluation of the proposed pipeline. In this section, we will provide few details on each of them. See Figure \ref{fig:datasets} for an example of the included plankton species. 

\paragraph{Lensless} The Lensless microscope dataset was introduced in \cite{Pastore2020}. It consists of images acquired using a lensless microscope, extracted from 1-minute videos. It includes 10 classes, with 640 color images each. Figure \ref{fig:lenslessdata} shows sample images for seven classes included in the dataset. 
\paragraph{WHOI 40} This dataset was released in \cite{Pastore2020} and it is a subset of the Woods Hole Oceanographic Institution (WHOI) Plankton Dataset\footnote{\url{https://hdl.handle.net/10.1575/1912/7341}} (years 2011-2014). See Figure \ref{fig:whoi} for an example of seven classes included in the dataset. 
Images were collected in situ by an automated submersible imaging-in-flow cytometry exploiting an Imaging FlowCytobot (IFCB).   
The WHOI 40 dataset includes 40 classes, each represented by 100 grayscale images, randomly selected from the original data. 
\paragraph{WHOI 22} This dataset was introduced in \cite{sosik2007automated}, and contains images extracted from the WHOI dataset. It includes 22 species, 300 grayscale images each. 

\subsection{Experiment Details}\label{subsec:exp_details}
In our experiments, the entire pipeline is implemented in PyTorch~\cite{NEURIPS2019_9015}.  Feature extraction is performed using DenseNet201, pre-trained on ImageNet without further fine-tunings. The network has approximately 20M parameters and achieves a top-5 accuracy of 0.9446 on the ImageNet validation set \cite{huang2018densely}. Input images are resized to $128\times128$ pixels and normalized accordingly to DenseNet201 specification. With our input format, DenseNet201 produces an output shape given by $(\text{channels, height, width})=(1920, 4, 4)$. The output is then reshaped to obtain higher performances with the convolutional (V)AE. We performed several tests, noticing better performances with a lower number of input channels and a higher width and height compared to the original output shape. Thus, we considered two candidate shapes: the first one with an approximate homogeneous distribution among channels, width and height $r_1=(30,32,32)$ and another one with a lower number of channels and a corresponding higher width $r_2=(3,32,320)$. We report in Section \ref{subsec:results} the results obtained with both shapes.
The AE and VAE used to produce our results are composed by an encoder made of three convolutional layers. The decoder is implemented with three convolutional transpose layers. The VAE has two additional dense layers after the encoder, to parametrize the mean and the log variance of the multivariate distributions in the latent space. AE and VAE are trained for a total of $100$ epochs using SGD and Adam optimizer respectively. Learning rate is initialized to 0.001 for both optimizers with an exponential learning rate decay for SGD. Batch size is set to 64. Hyperparameters were selected with a cross-validation procedure, adopting a k-fold approach (k=5).
We used the scikit-fuzzy~\cite{joshwarner20193541386} implementation of Fuzzy K-Means~\cite{fuzzy_k_means} for clustering the latent space data.
Results in Table~\ref{tab:sup_comp} are obtained using the sklearn implementation of Kernel ridge \cite{saunders1998ridge} and implementing a neural network classifier in TensorFlow~\cite{tensorflow2015-whitepaper}. The neural network is composed of 2 hidden dense layers of 256 and 128 neurons. It is trained for $100$ epochs, with batch size of $32$ and using stochastic gradient descent (SGD) optimizer with a learning rate of $0.01$. For Kernel ridge we used a Gaussian kernel. Hyper-parameters are tuned using grid search maximizing the classification accuracy.

\subsection{Results}\label{subsec:results}

We applied our pipeline to the three plankton datasets, described in Section \ref{subsec:datasets}. First, we verified the impact on the models' performances of the type of input used for training. To do this, we compared both AE and VAE trained either on the original images or on the pre-trained features reshaped as described in Section \ref{subsec:exp_details}. We performed multiple experiments considering five different latent space sizes, evaluating the results in terms of clustering purity and number of overlaps (see Section \ref{subsec:pipeline}) on the available test set, for each of the considered datasets. We specify that, since WHOI 40 is not originally distributed with a specific test set, we therefore performed a 80:20 train/test split as a preliminary step for this dataset. A purity of one with zero overlaps corresponds to clusters perfectly agreeing with the ground truth.

To prove the robustness of our method, we adopted a k-fold approach (k=5) repeating each experiment five times, with different train and validation splits. We then report mean and standard deviation for the purity and the number of overlaps on the test set for each dataset, in Tables~\ref{tab:z_lens}, \ref{tab:z_whoi}, \ref{tab:z_sosik}.

\begin{figure}[H]
\centering
\subfloat[Original images.]{\includegraphics[width=0.48\linewidth]{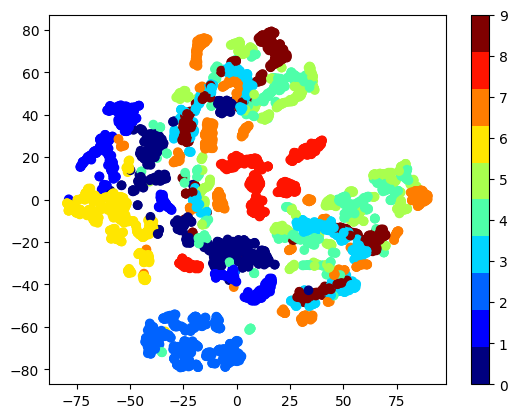}
\label{fig:z_lens_original}}
\hfil
\subfloat[Pre-trained features.]{\includegraphics[width=0.48\linewidth]{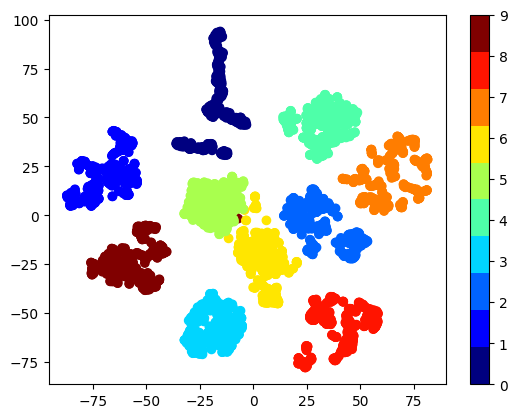}
\label{fig:z_lens_r2}}
\caption{t-SNE of the latent space learnt by a VAE on the lenseless dataset.} \label{fig:z_lens}
\end{figure}

Table \ref{tab:z_lens} shows our results on the Lensless dataset.  As we can observe, using pre-trained features significantly increases the purity with an average improvement of $30\%$ over the original images, considering both types of reshaping.
It is possible to appreciate such improvement in Figure \ref{fig:z_lens}, where an instance of learnt latent space, with and without, pre-trained features is shown. Thus, motivating the adoption of pre-trained features in our pipeline.  Moreover, using an input feature reshape with three channels $r_2=(3,32,320)$ results in a slight improvement over the case with more channels $r_1=(30,32,32)$. It is worth noticing that the VAE generally outperforms the AE for all the considered inputs and latent space sizes, with $Z=500$ giving an average test purity of $0.98$ without any overlap. 
\vspace*{-1em}
Table \ref{tab:z_whoi} and Table \ref{tab:z_sosik} show our results on the WHOI 40 dataset and the WHOI 22, respectively. 
These two datasets pose different challenges. The WHOI 40 includes a relatively high number of classes, with a more coarse granularity when compared to WHOI 22. Moreover, WHOI 22 includes images that contain multiple plankton cells, as well as a class for heterogeneous detritus (see, for instance, the last two images on the right in Figure \ref{fig:whoi}). In Table \ref{tab:z_lens}, \ref{tab:z_whoi} and \ref{tab:z_sosik}, the notation $FE_{r_{1/2}}$ refers to pre-trained features reshaped as $r_{1/2}$.
As we can observe, using an input feature reshape with three channels $r_2=(3,32,320)$ results now in a significant improvement for VAE over the case with more channels $r_1=(30,32,32)$. 
Generally, the best performances correspond to a latent space size $Z=500$, with a significant improvement with respect to small sizes.
The highest average test purity for the WHOI 40 corresponds to $0.77$ with four overlaps (i.e., four couples of ground truth species overlapping with the same clusters). The highest average test purity for the WHOI 22 is equal to $0.68$ with two overlaps.

We benchmarked our results using a state-of-the-art unsupervised learning pipeline based on a set of $131$ hand-crafted features and fuzzy k-means. The method was introduced in \cite{Pastore2020}, the same paper where two of the datasets used in this analysis (i.e., Lensless and WHOI 40) were released.
To our knowledge, a state-of-the-art clustering benchmark was not available for the WHOI 22. To benchmark our results, we used the approach described in \cite{Pastore2020} to perform clustering on the WHOI 22, based on the pipeline described in their paper. 
As shown in Table \ref{tab:baseline}, our best embedding (feature  reshape $r_2$, VAE with latent space size $Z=500$), outperforms the state-of-the-art approach for all the datasets included in our work. Our results are marginally better in terms of purity for the Lensless dataset. They are instead significantly better when considering the two more challenging datasets WHOI 40 and WHOI 22. We obtain not only a higher purity, but also a reduction in the number of overlaps. It is worth noticing that the purity and number of overlaps standard deviation among the different experimental runs is low for all the datasets, proving the robustness of the proposed methodology. 

\begin{table}[H]
	\centering
	\caption{
		Clustering purity on Lensless for latent space size $Z$.}
	\resizebox{\columnwidth}{!}{
	\huge
	\begin{tabular}{ l | l l l l l }
		\hline
		Algorithm/Z & 10 & 30 & 50 & 100 & 500\\\hline\hline
		image-AE & \begin{tabular}{@{}l@{}} $0.47 \pm 0.05$ \\ $(1.4 \pm 0.5)$\end{tabular} & \begin{tabular}{@{}l@{}}$0.53 \pm 0.03$ \\ $(1.4 \pm 0.49)$ \end{tabular} & \begin{tabular}{@{}l@{}}$0.56 \pm 0.02$ \\ $(1.6 \pm 0.49)$ \end{tabular}& \begin{tabular}{@{}l@{}}$0.55 \pm 0.01$ \\ $(1.8 \pm 0.4)$ \end{tabular}& \begin{tabular}{@{}l@{}}$0.60 \pm 0.01$ \\ $(1.4 \pm 0.49)$ \end{tabular}\\\hline
		
		image-VAE & \begin{tabular}{@{}l@{}} $0.53 \pm 0.017$ \\ $(1.4 \pm 0.5)$ \end{tabular} &  \begin{tabular}{@{}l@{}} $0.55 \pm 0.04$ \\$(1.6 \pm 0.49)$ \end{tabular} & \begin{tabular}{@{}l@{}} $0.58 \pm 0.01$ \\ $(2.0 \pm 0.63)$ \end{tabular} & \begin{tabular}{@{}l@{}} $0.59 \pm 0.01$ \\ $(1.6 \pm 0.48)$ \end{tabular} & \begin{tabular}{@{}l@{}} $0.62 \pm 0.01$ \\ $(2.0 \pm 0.0)$ \end{tabular} \\\hline
		
		$FE_{r_1}$-AE   & \begin{tabular}{@{}l@{}} $0.38 \pm 0.02$ \\ $(1.6 \pm 0.5)$ \end{tabular} & \begin{tabular}{@{}l@{}} $0.62 \pm 0.02$ \\$(1.2 \pm 0.40)$ \end{tabular} & \begin{tabular}{@{}l@{}} $0.75 \pm 0.04$ \\ $(0.4 \pm 0.49)$ \end{tabular} & \begin{tabular}{@{}l@{}} $0.84 \pm 0.03$ \\ $(0.2 \pm 0.4)$ \end{tabular} & \begin{tabular}{@{}l@{}} $0.95 \pm 0.01$ \\ $(0.0 \pm 0.0)$ \end{tabular} \\\hline
		
		$FE_{r_1}$-VAE & \begin{tabular}{@{}l@{}} $0.94 \pm 0.01$ \\ $(0.0 \pm 0.0)$ \end{tabular} & \begin{tabular}{@{}l@{}} $0.97 \pm 0.01$ \\$(0.0 \pm 0.0)$ \end{tabular} & \begin{tabular}{@{}l@{}} $0.97 \pm 0.01$ \\ $(0.0 \pm 0.0)$ \end{tabular} & \begin{tabular}{@{}l@{}} $0.97 \pm 0.01$ \\ $(0.0 \pm 0.0)$ \end{tabular} & \begin{tabular}{@{}l@{}} $0.97 \pm 0.002$ \\ $(0.0 \pm 0.0)$ \end{tabular} \\\hline
		
		$FE_{r_2}$-AE  &\begin{tabular}{@{}l@{}} $0.40 \pm 0.05$ \\ $(1.6 \pm 0.5)$ \end{tabular} & \begin{tabular}{@{}l@{}} $0.70 \pm 0.02$ \\$(0.8 \pm 0.4)$ \end{tabular} & \begin{tabular}{@{}l@{}}$0.75 \pm 0.03$ \\ $(0.4 \pm 0.49)$ \end{tabular} & \begin{tabular}{@{}l@{}} $0.86 \pm 0.05$ \\ $(0.2 \pm 0.4)$ \end{tabular} & \begin{tabular}{@{}l@{}} $0.97 \pm 0.01$ \\ $(0.0 \pm 0.0)$ \end{tabular}   \\\hline
		
		$FE_{r_2}$-VAE & \begin{tabular}{@{}l@{}} $0.98 \pm 0.01$ \\ $(0.0 \pm 0.0)$ \end{tabular} & \begin{tabular}{@{}l@{}} $0.98 \pm 0.03$ \\$(0.0 \pm 0.0)$ \end{tabular} & \begin{tabular}{@{}l@{}}$0.98 \pm 0.01$ \\ $(0.0 \pm 0.0)$ \end{tabular} & \begin{tabular}{@{}l@{}} $0.98 \pm 0.02$ \\ $(0.0 \pm 0.0)$ \end{tabular} & 
		\begin{tabular}{@{}l@{}} $0.98 \pm 0.02$ \\ $(0.0 \pm 0.0)$ \end{tabular} \\\hline
	\end{tabular} 
	}
	\label{tab:z_lens}
\end{table}

\vspace*{-1em}

\begin{table}[H]
	\centering
	\caption{
			Clustering purity on WHOI 40 for latent space size $Z$.
			}
	\resizebox{\columnwidth}{!}{
		\huge
	\begin{tabular}{ l | l l l l l }
		\hline
		Algorithm/Z & 10 & 30 & 50 & 100 & 500\\\hline\hline
		$FE_{r_1}$-AE   & 
		\begin{tabular}{@{}l@{}} $0.25 \pm 0.01$ \\ $(11.8 \pm 1.16)$ \end{tabular} &  \begin{tabular}{@{}l@{}} $0.41 \pm 0.03$ \\$(8.0 \pm 1.26)$ \end{tabular} &
		\begin{tabular}{@{}l@{}} $0.50 \pm 0.02$ \\ $(6.8 \pm 1.46)$ \end{tabular} &
		\begin{tabular}{@{}l@{}} $0.59 \pm 0.01$ \\ $(6.6 \pm 1.01)$ \end{tabular} &
		\begin{tabular}{@{}l@{}} $0.69 \pm 0.01$ \\ $(4.8 \pm 0.74)$ \end{tabular}\\\hline
		$FE_{r_1}$-VAE & 
		\begin{tabular}{@{}l@{}} $0.21 \pm 0.01$ \\ $(13.8 \pm 0.99)$ \end{tabular} & 
		\begin{tabular}{@{}l@{}} $0.22 \pm 0.02$ \\$(14.4 \pm 1.62)$ \end{tabular} &
		\begin{tabular}{@{}l@{}} $0.30 \pm 0.04$ \\ $(13.2 \pm 1.93)$ \end{tabular} &
		\begin{tabular}{@{}l@{}} $0.40 \pm 0.08$ \\ $(10.6 \pm 1.2)$ \end{tabular} & 
		\begin{tabular}{@{}l@{}} $0.67 \pm 0.04$ \\ $(5.4 \pm 1.35)$ \end{tabular}\\\hline
		$FE_{r_2}$-AE &
		\begin{tabular}{@{}l@{}} 	$0.25 \pm 0.01$ \\ $(13.2 \pm 0.74)$ \end{tabular} & 
		\begin{tabular}{@{}l@{}} $0.42 \pm 0.03$ \\$(9.0 \pm 0.63)$ \end{tabular} &
		\begin{tabular}{@{}l@{}}$0.51 \pm 0.01$ \\ $(6.2 \pm 0.40)$ \end{tabular} &
		\begin{tabular}{@{}l@{}} $0.62 \pm 0.01$ \\ $(5.8 \pm 1.16)$ \end{tabular} & 
		\begin{tabular}{@{}l@{}} $0.72 \pm 0.006$ \\ $(4.6 \pm 0.80)$ \end{tabular}\\\hline
		
		$FE_{r_2}$-VAE &
		\begin{tabular}{@{}l@{}} $0.66 \pm 0.01$ \\ $(5.8 \pm 0.75)$ \end{tabular} &
		\begin{tabular}{@{}l@{}} $0.71 \pm 0.02$ \\$(5.8 \pm 1.16)$ \end{tabular} &
		\begin{tabular}{@{}l@{}}$0.73 \pm 0.02$ \\ $(5.2 \pm 0.98)$ \end{tabular} &
		\begin{tabular}{@{}l@{}} $0.77 \pm 0.01$ \\ $(3.8 \pm 1.16)$ \end{tabular} &
		\begin{tabular}{@{}l@{}} $0.77 \pm 0.01$ \\ $(4.0 \pm 0.63)$ \end{tabular}\\\hline
	\end{tabular}
	}
	\label{tab:z_whoi}
\end{table}

\vspace*{-1em}

\begin{table}[H]
	\centering
	\caption{
		Clustering purity on WHOI 22 for latent space size $Z$.}
	\resizebox{\columnwidth}{!}{
		\huge

	\begin{tabular}{ l | l l l l l }
		\hline
		Algorithm/Z & 10 & 30 & 50 & 100 & 500\\\hline\hline
		$FE_{r_1}$-AE   & 
		\begin{tabular}{@{}l@{}} $0.22 \pm 0.02$ \\ $(5.0 \pm 1.1)$ \end{tabular} &
		\begin{tabular}{@{}l@{}} $0.40 \pm 0.02$ \\$(2.6 \pm 0.8)$ \end{tabular} &
		\begin{tabular}{@{}l@{}} $0.48 \pm 0.01$ \\ $(2.4 \pm 0.48)$ \end{tabular} &  
		\begin{tabular}{@{}l@{}} $0.55 \pm 0.01$ \\ $(1.0 \pm 0.0)$ \end{tabular} &
		\begin{tabular}{@{}l@{}} $0.65 \pm 0.02$ \\ $(1.4 \pm 0.49)$ \end{tabular}\\\hline
		$FE_{r_1}$-VAE & 
		\begin{tabular}{@{}l@{}} $0.22 \pm 0.02$ \\ $(7.6 \pm 1.2)$ \end{tabular} &
		\begin{tabular}{@{}l@{}} $0.30 \pm 0.01$ \\$(5.8 \pm 1.93)$ \end{tabular} & 
		\begin{tabular}{@{}l@{}} $0.33 \pm 0.07$ \\ $(5.8 \pm 1.32)$ \end{tabular} & 
		\begin{tabular}{@{}l@{}} $0.44 \pm 0.07$ \\ $(3.8 \pm 1.72)$ \end{tabular} & 
		\begin{tabular}{@{}l@{}} $0.68 \pm 0.01$ \\ $(2.0 \pm 0.4)$ \end{tabular}\\\hline
		$FE_{r_2}$-AE  &
		\begin{tabular}{@{}l@{}} $0.24 \pm 0.02$ \\ $(4.2 \pm 1.16)$ \end{tabular} & 
		\begin{tabular}{@{}l@{}} $0.38 \pm 0.01$ \\$(3.6 \pm 0.49)$ \end{tabular} &
		\begin{tabular}{@{}l@{}}$0.51 \pm 0.02$ \\ $(2.2 \pm 0.74)$ \end{tabular} &
		\begin{tabular}{@{}l@{}} $0.60 \pm 0.02$ \\ $(1.6 \pm 0.49)$ \end{tabular} & 
		\begin{tabular}{@{}l@{}} $0.66 \pm 0.01$ \\ $(1.2 \pm 0.4)$ \end{tabular}\\\hline
		$FE_{r_2}$-VAE &
		\begin{tabular}{@{}l@{}} $0.63 \pm 0.004$ \\ $(2.0 \pm 0.63)$ \end{tabular} & 
		\begin{tabular}{@{}l@{}} $0.66 \pm 0.01$ \\$(1.6 \pm 0.49)$ \end{tabular} &
		\begin{tabular}{@{}l@{}}$0.68 \pm 0.005$ \\ $(1.6 \pm 0.49)$ \end{tabular} &
		\begin{tabular}{@{}l@{}} $0.68 \pm 0.006$ \\ $(1.4 \pm 0.5)$ \end{tabular} &
		\begin{tabular}{@{}l@{}} $0.68 \pm 0.01$ \\ $(1.8 \pm 0.4)$ \end{tabular} \\\hline
	\end{tabular} 
	}
	\label{tab:z_sosik}
\end{table}

\vspace*{-1em}

\begin{table}[H]
	\centering
	\caption{
		Purity comparison between our best average results and the available state-of-the-art.}
	\begin{tabular}{ l | l l l l l }
		\hline
		Algorithm/Dataset & Lensless & WHOI 40 & WHOI 22 \\\hline\hline
	    Pipeline from \cite{Pastore2020}  & 0.93 (0) \cite{Pastore2020} & 0.71 (5) \cite{Pastore2020} & 0.56 (3)\\\hline
	    Ours  & \textbf{0.98 (0)} & \textbf{0.77 (4)} & \textbf{0.68 (2)} \\\hline

	\end{tabular} 
	\label{tab:baseline}
\end{table}

Finally, we benchmarked the proposed approach and the quality of our lower dimensional embeddings, with respect to supervised algorithms.
To this end, we considered two different classifiers i.e., a Fully Connected neural network (FC) and ridge regression (see Section \ref{subsec:exp_details}), trained on top of our best embedding for all three datasets. We performed five different experiments, comparing the resulting best test classification accuracy against available state-of-the-art supervised approaches, based on different types of classifiers. For the WHOI 22, we compared our results with the ones reported in  \cite{Zheng2017} and \cite{sosik2007automated}, where a set of hand-crafted features, appositely designed and selected for plankton images is engineered, and fed to a SVM classifier. For the WHOI 40 dataset and the Lensless dataset, we compared our results with the ones reported in  \cite{Pastore2020}, where a set of 131 hand-crafted features is extracted and fed to a two layers neural network (for Lensless) and a Random Forest (RF) classifier (for WHOI-40).  
As we can observe in Table \ref{tab:sup_comp}, a ridge regression classifier on top of our best embedding outperforms the state-of-the-art supervised classification results, for all the three datasets included in our analysis (1.000 versus 0.980, 0.957 versus 0.790 and 0.883 versus 0.880, for the Lensless, the WHOI 40 and the WHOI 22, respectively).
 These results proved that our embedding provides a better representation of the input data compared to hand-crafted features, specifically designed for plankton cells. 
For completeness, in Table~\ref{tab:pipe_time}, we indicated the total time required to execute  our pipeline for the best model. 

\begin{table}[H]
	\centering
	\caption{
		Supervised learning benchmarks.
		The state-of-the-art results are directly taken from the original publications.  
	}
	\begin{tabular}{ l | l l l}
		\hline
		Algorithm/Dataset & Lensless & WHOI 40 & WHOI 22 \\\hline\hline
		Our embedding + FC & 1.000 & 0.948 & 0.868 \\\hline
		Our embedding + ridge & \textbf{1.000} & \textbf{0.957} & \textbf{0.883}  \\\hline
        Features from \cite{Pastore2020} + FC & 0.980\cite{Pastore2020} & --- & ---\\\hline
		Features from \cite{Pastore2020} + RF & --- & 0.790\cite{Pastore2020} & ---\\\hline
		Features from \cite{Zheng2017} + SVM & --- & ---  & 0.880\cite{Zheng2017}\\\hline
	    Features from \cite{sosik2007automated} + SVM  & --- & --- & 0.880\cite{sosik2007automated}\\\hline
	\end{tabular} 
	\label{tab:sup_comp}
\end{table}

\vspace*{-1em}

\begin{table}[H]
	\centering
	\caption{
		Time requested for the execution of our pipeline (best model).}
	\begin{tabular}{ l l l }
		\hline

		Lensless & WHOI 40 & WHOI 22 \\\hline\hline
		$(238.56 \pm 2.00)$ s & $(195.75 \pm 1.04)$ s & $(194.93 \pm 2.51)$ s  \\\hline
	\end{tabular}
	\label{tab:pipe_time}
\end{table}

\section{Conclusions}\label{sec:conclusions}
In this paper, we introduced an efficient unsupervised learning pipeline for the characterization of plankton images. Input images are pre-processed and fed to a neural network (DenseNet201 in our experiments) pre-trained on ImageNet, without fine-tuning. The resulting set of features are used as inputs to train an encoder-decoder neural network (an AE or a VAE) and the resulting latent space representations of the inputs are used as a lower dimensional set of embedded features, that are then passed to a clustering algorithm (a fuzzy k-means in our experiments).
We exploited three datasets extracted from two different acquisition systems and posing different challenges.  
The Lensless dataset, with a coarse granularity but characterized by low-resolution and noisy images; the WHOI 40, less coarse with respect to the Lensless and with a relatively higher number of classes (40) and the WHOI 22, with fine-grained features. We showed that a VAE with latent space size $Z = 500$ and pre-trained input features, reshaped as $\text{(channels,height,width)}=(3,32,320)$, generally gives the best results in terms of purity and number of overlaps, for all the datasets included in our work.
We further showed that our approach outperforms state-of-the-art unsupervised learning approaches \cite{Pastore2020} where hand-crafted features are engineered and used for clustering. 
We further proved the quality of the embedded features produced by our pipeline using a supervised classification framework (in terms of test accuracy). Precisely, we showed that our embedding features coupled to a ridge regression classifier outperforms state-of-the-art classifiers where hand-crafted features are used as input for SVM \cite{sosik2007automated,Zheng2017}, fully connected neural networks and random forests \cite{Pastore2020}.
The VAE architecture considered for the tests of the proposed pipeline is shallow (see Section \ref{subsec:exp_details}), and the ImageNet pre-trained DenseNet201 is only used for features extraction. Hence, the pipeline can be run on embedded devices (e.g., a Rasperry-Pi), allowing for in situ recognition, which may be fundamental for plankton population studies.
It is worth underlining that our pipeline is general with respect to the source of input data. A complete analysis of the performances on other kind of data is out of the scope of this paper, however, the proposed pipeline can be potentially extended to other domains.
A significant advantage of our approach is indeed represented by the usage of pre-trained features and an encoder-decoder network to obtain a quality embedding for plankton data. Differently from the hand-crafted features exploited in the works we used as benchmarks, these features do not require any engineering nor tuning at any step to adapt to a specific dataset. For instance, the computation of shape-descriptors is instead a multi step-process, starting with a segmentation algorithm (as done in \cite{sosik2007automated,Pastore2020}) to identify the plankton cell. The segmentation can be performed using image processing tools or deep learning. The quality of the features is then highly dependent on the quality of the segmentation, which requires tuning according to specific properties of the dataset (e.g., acquisition system, brightness, noise). Avoiding this step, our pipeline represents an efficient approach with a significant advantage in terms of time and resources.
Finally, as a further development, the implementation of an end-to-end solution would be crucial for an easy deployment in real-life scenarios. Additionally, it would be interesting and useful to test the approach for anomaly detection. These aspects are currently under study.

\section*{Acknowledgment}
P.A., M.L., M.R. and L.R. acknowledge the financial support of the European Research Council (grant SLING 819789). L.R. acknowledges the financial support of the AFOSR projects FA9550-18-1-7009, FA9550-17-1-0390 and BAA-AFRL-AFOSR-2016-0007 (European Office of Aerospace Research and Development), and the EU H2020-MSCA-RISE project NoMADS-DLV-777826. We gratefully acknowledge the support of NVIDIA Corporation for the donation of the Titan Xp GPUs and the Tesla k40 GPU used for this research.

\newpage


\bibliographystyle{IEEEtran}

\bibliography{bibliography}

\end{document}